\documentclass{article}
\usepackage[preprint]{neurips}
 
\usepackage[utf8]{inputenc}  
\usepackage[T1]{fontenc}   

\usepackage{todonotes}

\usepackage{titletoc}
\usepackage[page, header, toc, page]{appendix}

\usepackage{hyperref}    
\usepackage{url}        
\usepackage{booktabs}    
\usepackage{amsfonts}  
\usepackage{verbatim}
\usepackage{nicefrac} 
\usepackage{microtype}     
\usepackage{xcolor}    
\usepackage{amsmath,amssymb,amsthm}
\usepackage{mdframed}
\usepackage{algorithm}
\usepackage{algpseudocode}
\usepackage{enumitem} 
\usepackage{natbib}
\usepackage{etoolbox}

\usepackage{xcolor,graphicx}
\usepackage{colortbl}
\usepackage{tcolorbox}

\usepackage{wrapfig} 
\usepackage{subcaption}
\usepackage{booktabs}
\algnewcommand{\States}[1]{%
\Statex \hspace{-0.5em} \hspace{-\algorithmicindent}\textbf{States:} #1%
} 

\usepackage{multicol}
\usepackage{multirow}

\setlist{nolistsep}
\setlist{leftmargin=*} 

\definecolor{darkgreen}{rgb}{0,0.3,0}
\definecolor{darkblue}{rgb}{0.0,0.0,0.65}
\definecolor{darkred}{rgb}{0.3,0.0,0.0}
\hypersetup{
colorlinks = true,
citecolor  = darkblue,
linkcolor  = darkred,
filecolor  = darkblue,
urlcolor   = darkblue,
}

\usepackage{mathtools}
\usepackage{thmtools}

\newcommand{\algcomment}[1]{\textcolor{blue!70!black}{\hfill
\small{$\triangleright$\texttt{\hspace{2pt}#1}}}}

\definecolor{lightgray}{gray}{0.6}
\definecolor{DarkGreen}{rgb}{0, 0.5, 0}

\newcommand{\norm}[1]{\|#1\|}

\newcommand{\R}{\mathbb{R}}

\usepackage{soul} 

\title{Dion2: A Simple Method to Shrink Matrix in Muon}

\author{
Kwangjun Ahn\textsuperscript{1} \quad Noah Amsel\textsuperscript{2} \quad John Langford\textsuperscript{1}\\\\
\textsuperscript{1}Microsoft Research, AI Frontiers \\
  \textsuperscript{2}NYU  
}

\begin{document}

\maketitle

\begin{abstract}
The Muon optimizer enjoys strong empirical performance and theoretical grounding. However, the superlinear cost of its orthonormalization step introduces increasing overhead with scale. To alleviate this cost, several works have attempted to reduce the size of the matrix entering the orthonormalization step.
We introduce \textbf{Dion2}, a much simpler method for shrinking the matrix involved in Muon’s computation compared to prior approaches. At a high level, Dion2 selects a fraction of rows or columns at each iteration and orthonormalizes only those. This  sampling procedure makes the update \emph{sparse}, reducing both computation and communication costs which in turn improves the scalability of Muon.
\end{abstract}

\footnotesize{A preliminary Pytorch FSDP2 implementation is available at: \url{https://github.com/microsoft/dion/}}

\begin{figure}[h]
\centering  
\includegraphics[width=\linewidth]{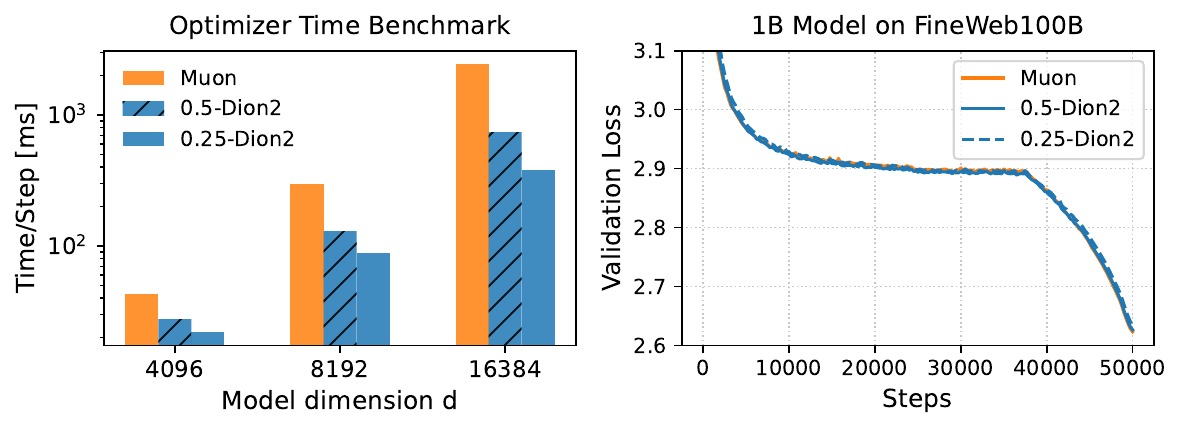} 
\caption{\footnotesize
We propose a simple method to reduce the size of the matrix entering Muon’s Newton-Schulz iterations while preserving Muon’s high update quality. 
\textbf{Left:} Shrinking the matrix leads to faster time per step (compute-only benchmark).
\textbf{Right:} Even orthonormalizing only 25\% of the matrix maintains update quality close to full Muon at the 1B-model / 100B-token training scale (final losses: Muon 2.623 vs.\ $0.25$-Dion2 2.635).
}
\label{fig:main}
\end{figure}

\section{Introduction}
\label{sec:intro}
Training state-of-the-art AI models requires millions of GPU-hours, so any improvement to the optimizer can significantly reduce computational cost.
Since its introduction, Adam(-W)~\citep{kingma2014adam,loshchilov2018decoupled} has remained the default choice for large-scale training.
A decade later, \textbf{Muon}~\citep{jordan2024muon} has emerged as a compelling alternative, offering superior optimization efficiency~\citep{liu2025muon} and improved large-batch scaling~\citep{shah2025practical}.
Notably, Muon has already been adopted in frontier models~\citep{kimi2025k2,zeng2025glm}.

The theoretical motivation for Muon is to control the growth of each layer's activations during training~\citep{bernstein2025deriving}. Its rule for updating the weight matrices follows immediately from this principle. Besides its theoretical motivation,  empirical advantages of Muon include robust hyperparameter transfer across scales~\citep{bernstein2024modular,large2024scalable,pethick2025training}, improved training stability~\citep{liu2025muon,kimi2025k2}, improved large-batch scaling~\citep{shah2025practical} and faster training~\citep{liu2025muon}.

To illustrate, consider a linear layer $y = W x$ with $x \in \R^{\texttt{fan-in}}$ and $y \in \R^{\texttt{fan-out}}$.
The change in activations due to a weight update satisfies
\begin{align*}
\norm{\Delta y}_{\mathrm{RMS}}
= \norm{\Delta W x}_{\mathrm{RMS}}
\le \norm{\Delta W}_{\mathrm{RMS}\to\mathrm{RMS}}\cdot\norm{x}_{\mathrm{RMS}}
= \norm{\Delta W}_{\mathrm{RMS}\to\mathrm{RMS}},
\end{align*}
assuming the input $x$ has unit RMS norm.
To guarantee $\norm{\Delta y}_{\mathrm{RMS}} \le \eta$, it suffices for the update to take the form
\begin{align} \label{orth-update}
\Delta W = \eta \sqrt{\tfrac{\texttt{fan-out}}{\texttt{fan-in}}}\cdot O,
\quad \text{where $O$ has unit spectral norm.}
\end{align}
Under this condition, $\norm{\Delta W}_{\mathrm{RMS}\to\mathrm{RMS}} = \eta$ follows immediately.

In Muon, condition~\eqref{orth-update} is satisfied by setting $O$ to  the projection of the momentum matrix onto the set of orthonormal matrices.
Importantly, this orthonormalization step can be performed efficiently on GPUs via Newton-Schulz iterations, which require only matrix multiplications and additions~\citep{bernstein2024modular,jordan2024muon}.
Formally, the update takes the form $O_{\mathrm{Muon}} = \texttt{Newton–Schulz}(M)$ where $M$ is the momentum matrix. 
This choice ensures bounded activation changes and maximizes the quality of each update.

However, as detailed in \autoref{sec:scalability}, fully orthonormalizing the momentum matrix of every layer at every iteration can pose a computational and communication bottleneck at large scales \citep{essential2025muon}. In contrast, Adam does not have this overhead because its updates are element-wise. This raises a question:
\begin{center}
   \emph{Can we reduce the overhead of Muon's orthonormalization step?}
\end{center}
This question was first explored in \textsc{Dion} \citep{ahn2025dion}, which orthonormalizes a low-rank approximation of the momentum matrix. However, their low-rank approximation procedure, based on power iteration, introduces additional complexity.

In this work, we propose \textsc{Dion2}, a drastically simpler alternative. Instead of forming a low-rank approximation using power iteration, \textsc{Dion2} \textbf{directly selects a subset of rows (or columns) at each iteration and orthonormalizes only those}. As a result, the update is \emph{sparse}: only the selected rows (or columns) of each weight matrix are modified. In \autoref{sec:dion2}, we explain how we implement this approach and how its design builds on insights from prior work~\citep{ahn2025dion,modoranu2025fft}. Running Newton–Schulz on a reduced submatrix substantially lowers both compute costs and communication overhead in distributed settings.

As shown in \autoref{fig:main}, for a 1B model trained on 100B tokens of the FineWeb dataset,  updating just 25\% of the columns at each iteration nearly matches the quality of full Muon orthonormalization.
\autoref{fig:main} also shows that our method achieves substantial reductions in the wall-clock time of the orthogonalization procedure.

\section{Scalability of Muon}
\label{sec:scalability}

In this section, we provide a detailed discussion of Muon’s scalability properties. As noted above, whereas Adam relies on inexpensive element-wise operations, Muon requires matrix orthonormalization, which has super-linear complexity. This orthonormalization step is substantially more expensive than Adam’s element-wise updates, and its cost grows quickly with model size. Although sparse MoE architectures keep most matrices smaller, they also reduce overall model FLOPs, which increases the relative cost of the Newton–Schulz step \citep{essential2025muon}. Moreover, sparse architectures still include dense layers \citep{liu2024deepseek}, so large matrices remain.

There are additional challenges when weights are sharded, as is standard in distributed training. In this setting, the Newton–Schulz computation is difficult to parallelize evenly across devices. \citet{essential2025muon} analyze this overhead and propose using all-to-all communication along the weight-parallel dimension to de-duplicate Muon computation. This strategy is implemented in PyTorch FSDP2 by \citet{ahn2025dion}, and its compute–communication overlap characteristics are examined in \citet{lim2025motif2}. While this approach makes the overhead manageable in some settings, it does not fully resolve the scalability limitations.

Muon has been scaled successfully by Moonshot AI \citep{liu2025muon, kimi2025k2}. Their success was enabled by an alignment of several factors:
\begin{enumerate}[nosep]
\item As noted by \citet{liu2025proof}, earlier releases of PyTorch and Megatron-LM used DP-sharding strategies for optimizer states that were, by chance, favorable for Muon. Model and optimizer states were stored in large contiguous flat buffers, and data-parallel shards were produced by splitting this buffer. As a result, only tensors crossing a DP boundary required an additional gather. These advantageous strategies, however, have been deprecated in more recent releases.
\item They adopt a fine-grained MoE architecture with only a single dense layer (even fewer than in DeepSeek-V3 \citep{liu2024deepseek}), so most matrices remain small even at the one-trillion-parameter scale, keeping the Newton–Schulz cost manageable.
\item Their main distributed training strategy combines pipeline parallelism with expert parallelism, which naturally distributes Muon’s computation across devices with minimal communication.

\end{enumerate}

In sum, Muon has been successfully scaled, but its success relies on a subtle alignment of architectural, parallelism, and framework factors.  
For Muon to serve as a more general replacement for Adam, it benefits from additional scalability levers that relax these constraints. As \citet{essential2025muon} point out, one promising approach is to reduce the size of the matrix entering the Newton–Schulz iterations. 
This forms the main focus of the remainder of this work.

\newcommand{\fs}{\mathsf{X}}
\newcommand{\Topk}{\textsc{Top}_k} 
\newcommand{\I}{\mathcal{I}}
\newcommand{\J}{\mathcal{J}}
\newcommand{\K}{\mathcal{K}}
\newcommand{\Select}{\texttt{Select}}

\section{Dion2: A Simplest Recipe to Shrink the Muon Matrix}
\label{sec:dion2}

In this section, we introduce \textsc{Dion2}, a simple and effective method for reducing the size of the matrix that enters the Newton–Schulz iterations in Muon.
The pseudocode is presented in \autoref{alg:dion2}.
In words, \textsc{Dion2}
\begin{enumerate}
\item Selects an $\alpha$-fraction of the rows (or columns) of the momentum matrix.
\item Orthogonalizes the corresponding submatrix using Newton-Schulz.
\item  Updates the selected rows (or columns) of the weight matrix using the orthogonalized submatrix, as in \eqref{orth-update}.
\item Decays the selected rows (or columns) of the momentum matrix by a multiplicative factor $\mu$.
\end{enumerate}

In the remainder of this section, we give a detailed breakdown of the procedure, explain its advantages, and provide intuition for how it simplifies prior approaches.

\subsection{Algorithmic Components}

One key component of the algorithm is the selection method $\Select_\alpha(M)$. 
In our experiments, we evaluate two simple strategies: (i) selecting the subset of rows or columns that have the largest $\ell_1$ norm, and (ii) selecting a subset of rows or columns uniformly at random. Both perform well at the scales we tested; see \autoref{sec:experiments} for details. Given our results, it is likely that other selection strategies could perform equally well or even better. We leave further exploration to future work.

The second key component of our method is the selective decay mechanism, in which only the currently selected portion of the momentum is decayed:
\begin{align} \label{error-feedback}
M[\K, :] \gets \mu \cdot M[\K, :].
\end{align}
(In contrast, standard momentum does $M \gets \mu \cdot M$.) As shown in \autoref{sec:error-feedback} via ablation studies, this procedure is essential for the method to work.
This mechanism is analogous to \textsc{Dion}'s error-feedback approach, but adapted to our simpler submatrix selection scheme (see \autoref{sec:previous}).

\begin{algorithm}[t]
\caption{\textsc{Dion2} with fraction $\alpha$ on a matrix parameter $W$}\label{alg:dion2}
\begin{algorithmic}[1]
\Statex{\textbf{Hyperparameter}: Momentum decay factor $\mu =0.95$.}
\Statex  Row-selection version (column-selection is analogous): 
\Function{ $\alpha$-Dion2}{$G,~M$}
\State $M \gets M + G$ \algcomment{accumulate gradient into momentum} 
\State $\K \gets \Select_\alpha(M)$ \algcomment{select $\alpha$-fraction of rows}
 
\State $O \gets \texttt{NewtonSchulz}(M[\K, :])$  \algcomment{orthonormalize only the submatrix}
 
\State $M[\K, :] \gets \mu \cdot M[\K, :]$ \algcomment{in-place decay on selected rows}
\State $W[\K, :] \gets W[\K, :] -  \eta \sqrt{\tfrac{\texttt{fan-out}}{\texttt{fan-in}}}\cdot O$ \algcomment{sparse parameter update}
\EndFunction
\end{algorithmic}
\vspace{-0.8em}
\rule{\textwidth}{0.4pt}
\vspace{-1.3em}
\begin{algorithmic}
\State \textbf{Selection method.} We test two candidates for $\Select_\alpha(M)$:
\begin{enumerate}
\item (\texttt{Default}) selecting  the top rows/columns based on their $\ell_1$ norms, or
\item (\texttt{Random}) selecting uniformly at random.
\end{enumerate} 
\end{algorithmic}
\end{algorithm}

\subsection{Advantages of \textsc{Dion2}}
\label{sec:advantage}
Orthonormalizing only a subset of rows or columns directly mitigates the scalability issues  of Muon discussed in \autoref{sec:scalability}. Running Newton–Schulz on a submatrix reduces the amount of matrix arithmetic and therefore yields speedups in compute-bound regimes. As shown in \autoref{fig:main}, this reduction produces measurable wall-clock improvements. Because only the selected subset of rows needs to be reconstructed and communicated across devices, \textsc{Dion2} also reduces the communication volume.

There is a further advantage at even larger scales, when data parallelism spans multiple pods over data-center networking (DCN) or when hybrid sharding is used \citep{zhao2023pytorch}. A common deployment is to run model parallelism inside an inter-chip interconnects (ICI) domain and pure data parallelism across pods \citep{scaling-book}. Since DCN bandwidth is typically smaller than ICI bandwidth, shrinking the size of data-parallel (DP) synchronization is particularly valuable.

In these settings, \textsc{Dion2} supports a compressed DP-sync strategy analogous to \textsc{Dion}'s approach \citep[\S 3.3]{ahn2025dion}, provided the selection method does not require full synchronization of the entire momentum state. For example, with uniform random selection, only the selected submatrix $M[\K, :]$ needs to be synchronized across data-parallel workers. This causes the momentum states to diverge across replicas, but crucially, the information that \emph{is} synchronized suffices to compute the correct parameter update, just as full DP-sync would.

Hence, \textsc{Dion2} reduces DP-communication without sacrificing the  quality of the global weight update.  Efficient selection of the top columns by $\ell_1$ norm is more challenging to parallelize, but given the almost equal performance of random selection, approximations would likely work well (\emph{e.g.}, choosing the top columns from each datacenter instead of the top columns overall).

\subsection{Comparison with Prior Approaches}
\label{sec:previous}

Here, we explain how \textsc{Dion2} draws inspiration from two prior approaches—\textsc{Dion} \citep{ahn2025dion} and \textsc{Trion} \citep{modoranu2025fft}—while substantially simplifying them.

As noted in \autoref{sec:intro}, the problem of reducing the cost of orthonormalization was first explored in \textsc{Dion}. \textsc{Dion} constructs a low-rank approximation of the momentum matrix and orthonormalizes only this smaller matrix.
 If $M \approx MV \cdot V^\top$ is a low-rank approximation of the momentum matrix with $V^\top V = I$, then
\begin{equation}
\texttt{NewtonSchulz}(M) \approx \texttt{NewtonSchulz}(MVV^\top) = \texttt{NewtonSchulz}(MV) V^\top.
\end{equation}
Because the dimensions of $MV$ are much smaller than those of $M$, this approximation greatly reduces the cost of the Newton-Schulz orthogonalization step.
In \textsc{Dion}, $V$ is found by an amortized power-iteration procedure.

The key mechanism enabling this approximation is \textbf{error feedback}. The usual rule for updating the momentum matrix is $M \gets M + G;\, M \gets \mu M$, which ensures that $M$ is an exponential moving average of the gradients. Error feedback modifies this rule by decaying only the component of $M$ that was captured by the low-rank approximation:
\begin{equation}\label{error-feedback-original}
M \gets M + G; \quad M \gets M - (1 - \mu)MVV^\top \end{equation}
Intuitively, this update rule ensures that subsequent iterations focus on residual components not captured by the current low-rank approximation.
When the approximation is exact ($M = MVV^\top$), we recover standard momentum.

\textsc{Trion} \citep{modoranu2025fft} adopts the main ideas of \textsc{Dion}, but changes how the low rank approximation is computed. Instead of amortized power iteration, it uses an approximation based on the discrete cosine transform. Surprisingly, this simpler approximation performs even better than \textsc{Dion}, suggesting that finding a \emph{good} low-rank approximation is not essential. What truly matters is the error-feedback mechanism, which compensates even for large differences between $M$ and $MVV^\top$.  

Our method, \textsc{Dion2}, pushes this simplification further. We choose the simplest low-rank approximation: we simply select a fraction of the rows or columns and set the rest to zero. Treating this as a low rank approximation, the error-feedback update in \eqref{error-feedback-original} reduces to decaying only the selected rows or columns as in \eqref{error-feedback}.

We also note an alternative to low-rank approximation methods: \emph{block orthonormalization} \citep{boreiko2025towards, khaled2025muonbpscale}.
In this approach, the matrix is partitioned into blocks, each typically corresponding to a shard in sharded-weight settings, and each block is orthonormalized independently.
In particular, \textsc{MuonBP}~\citep{khaled2025muonbpscale} shows that periodically alternating block-wise steps with full-matrix orthonormalization yields a highly effective method.

\section{Experiments}
\label{sec:experiments}

In this section, we present experiments evaluating \textsc{Dion2}. 
Our experiments aim to demonstrate that:
\begin{enumerate}
    \item \textsc{Dion2} achieves better update quality than \textsc{Dion}, despite being much simpler.
    \item Surprisingly, the random selection method is also highly effective.
    \item Error feedback is the key mechanism enabling \textsc{Dion2}'s success. 
\end{enumerate}

\textbf{Experimental details.}  
We use the \textsc{Dion} codebase\footnote{\url{https://github.com/microsoft/dion/}} for our experiments. All experiments are conducted on the FineWeb dataset \citep{penedo2024fine}. The model configuration is summarized in \autoref{tab:scale_cfg}.  
For all experiments, we select the submatrix along the shorter dimension of the momentum matrix.
We use the same learning rate of $0.02$ for both Muon and Dion variants. This value has been shown to work well in the \textsc{Dion} codebase \citep{ahn2025dion}, and our goal is to demonstrate that \textsc{Dion2} performs well out-of-the-box with this learning rate. We adopt a constant learning rate schedule with a $25\%$ decay period.

\begin{table}[h]
\centering\small
\begin{tabular}{@{}ccccccc@{}}
\toprule
Total param   & $d_\text{model}$ / Layers / Heads & Seqlen &Batch Size token & Total Steps & Total Tokens  \\
\midrule
304M & 1024 / 16 / 8 & 2048 & 2.0~M & 10~K  & 20 B \\
1.01B  & 2048 / 16 / 16 & 2048 & 2.0~M & 50~K & 100 B \\ 
\bottomrule
\end{tabular}
\vspace{4pt}
\caption{\footnotesize Configurations for models in our experiments. }
\label{tab:scale_cfg}
\end{table}

\subsection{300M Model Experiments}

\begin{figure}[h]
\centering  
\includegraphics[width=\linewidth]{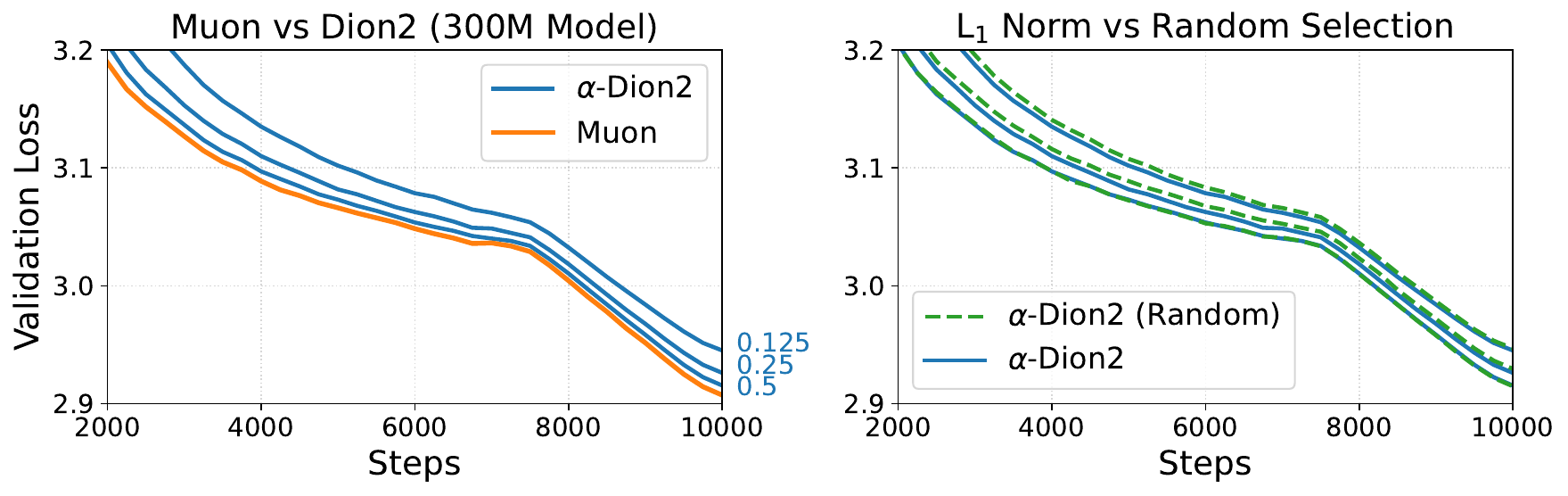} 
\caption{\footnotesize {\bf 300M model trained on 20B FineWeb.} \textbf{Left:} comparison of Muon and \textsc{Dion2} implemented with $\ell_1$-norm selection. \textbf{Right:} comparison of different selection methods in \textsc{Dion2}. Final losses are nearly identical. For the $\ell_1$-norm vs.\ random comparison:
($\alpha = 0.5$) $2.9154$ vs.\ $2.9148$,
($\alpha 0.25$) $2.9262$ vs.\ $2.9296$,
($\alpha = 0.125$) $2.9452$ vs.\ $2.9469$.}
\label{fig:dion_300m}
\label{fig:dion_300m}
\end{figure}

\begin{figure}[h]
\centering  
\includegraphics[width=\linewidth]{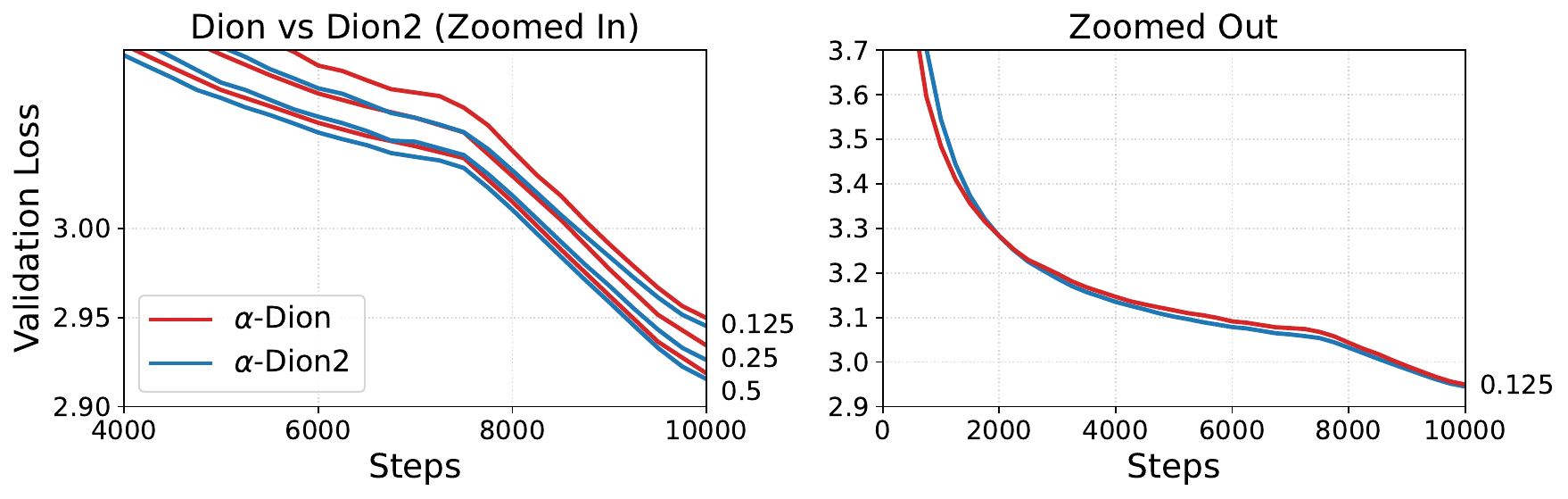} 
\caption{\footnotesize {\bf Dion vs. \textsc{Dion2} ($\ell_1$ selection) on the 300M model.} \textbf{Left:} zoomed-in view of validation loss. Right: zoomed-out view. \textsc{Dion2} achieves a better trade-off in update quality. Notably, it initially lags behind Dion but eventually catches up and surpasses it.}
\label{fig:dion_1v2}
\end{figure}

We first test \textsc{Dion2} on the 304M model from \autoref{tab:scale_cfg}. In the first experiment (\autoref{fig:dion_300m}, left), we vary the fraction of the submatrix used ($\alpha = 0.5, 0.25, 0.125$), selecting the top-$\alpha$ rows or columns based on their $\ell_1$ norm. As expected, reducing the fraction slightly decreases update quality, leading to slower convergence.

As shown in the right plot of \autoref{fig:dion_300m}, switching the selection method from $\ell_1$-norm–based to uniform random results in nearly identical convergence, suggesting that even very simple random selection is highly effective.

Next, we compare \textsc{Dion2} against \textsc{Dion} in \autoref{fig:dion_1v2}.
Despite its simpler design, \textsc{Dion2} achieves faster convergence. Interestingly, as shown in the plot of \autoref{fig:dion_1v2}, \textsc{Dion2} initially lags behind \textsc{Dion} but eventually catches up and surpasses it, ultimately achieving a lower final loss.

\subsection{Error-Feedback Ablations}
\label{sec:error-feedback}

\begin{figure}[h]
\centering  
\includegraphics[width=0.5\linewidth]{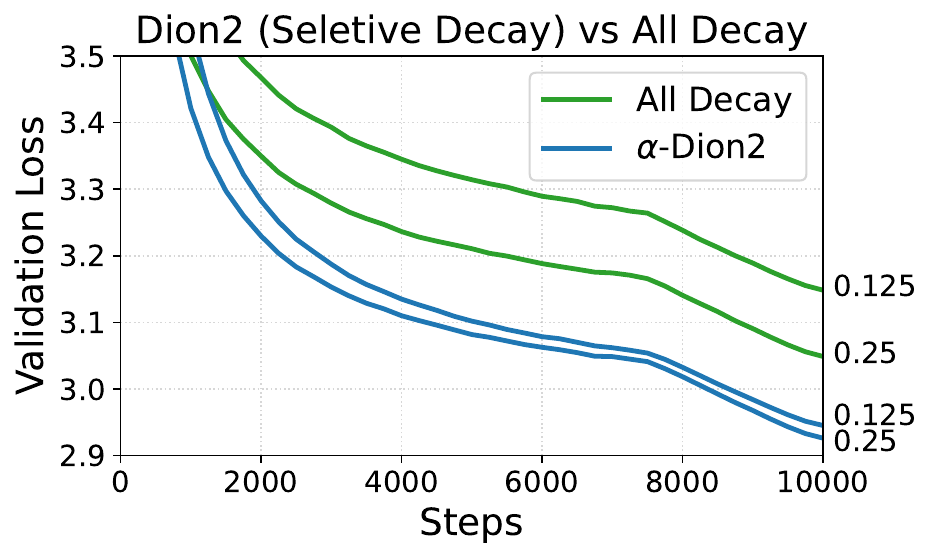} 
\caption{\footnotesize {\bf Error-Feedback Ablation.} \textsc{Dion2} decays only the selected rows or columns of the momentum matrix. We ablate this component by comparing against a variant that decays all rows/columns. The results show that the selective decay mechanism is critical for performance.}
\label{fig:error}
\end{figure}

We next evaluate the necessity of the error-feedback mechanism in \eqref{error-feedback}. Specifically, in \autoref{fig:error}, we compare \textsc{Dion2} against a variant that applies decay to the entire momentum matrix after the update:
\begin{align*} 
\textbf{Ablation:}\quad M \gets \mu \cdot M \quad \text{instead of} \quad M[\K, :] \gets \mu \cdot M[\K, :].
\end{align*}
As shown in \autoref{fig:error}, this variant leads to significantly higher loss, indicating that selective decay of the chosen submatrix is a crucial mechanism enabling effective reduction of the matrix size during orthonormalization.

\subsection{1B Model Experiments}

\begin{figure}[h]
\centering  
\includegraphics[width=\linewidth]{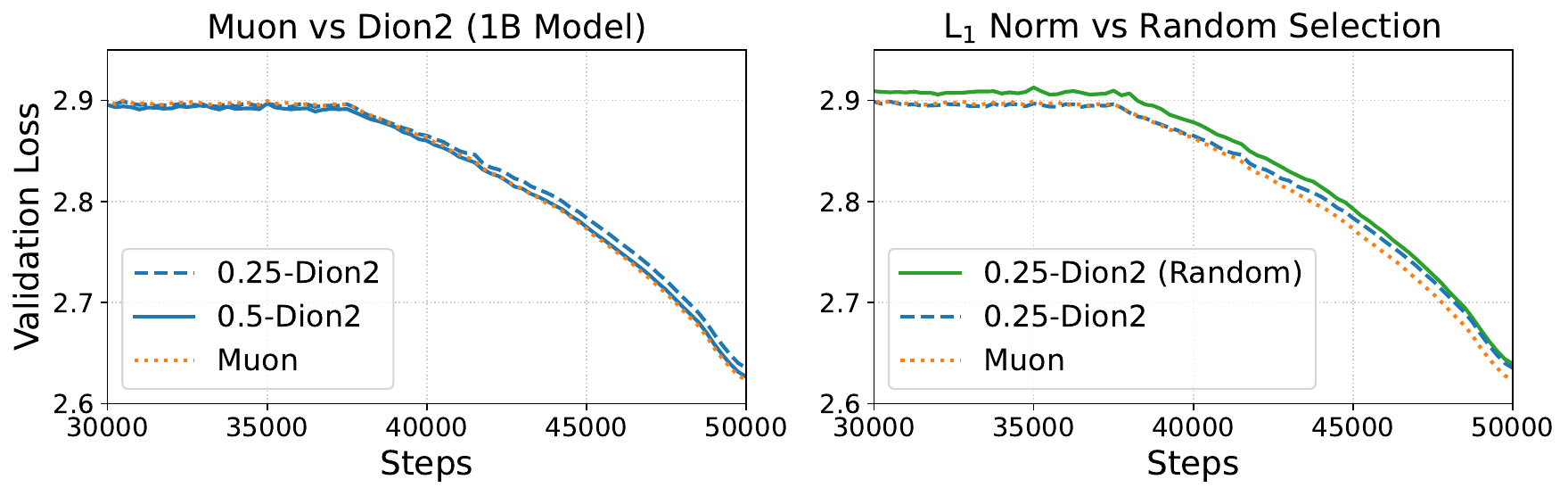} 
\caption{\footnotesize {\bf 1B model trained on 100B FineWeb.} Left: comparison of Muon and \textsc{Dion2}. Right: comparison of different selection methods in \textsc{Dion2}.}
\label{fig:dion_1b}
\end{figure}

We now present results for the 1B-parameter model from \autoref{tab:scale_cfg}. We first test the $\ell_1$-norm  based selection method. As shown in \autoref{fig:main} and \autoref{fig:dion_1b}, the results are very encouraging. Compared with the 300M model, the performance gap between 0.25-\textsc{Dion2} and Muon narrows indicating favorable scaling. In particular, in the zoomed-in plot (\autoref{fig:dion_1b}, left), 0.25-\textsc{Dion2} achieves almost identical validation loss to Muon before the learning rate decay and shows only a slight degradation afterward. The final validation losses are 2.623 for Muon and 2.635 for 0.25-\textsc{Dion2}.

In the right plot of \autoref{fig:dion_1b}, we evaluate \textsc{Dion2} using random selection. As the plot shows, the final validation losses after the learning rate decay are nearly identical to the $\ell_1$-norm based selection, indicating that the random selection method remains effective even at the 1B-parameter scale.

\subsection{Micro Benchmark}

As discussed in \autoref{sec:advantage}, reducing the size of the matrix that enters the orthonormalization step can lower both computation and communication costs. The magnitude of these benefits depends heavily on the specific training infrastructure, so it is difficult to capture them fully in a single experiment. A precise measurement would require running on a large-scale production training setup.

Nevertheless, we provide a preliminary micro benchmark to illustrate at least the computational savings. We initialize a simple 4-layer GPT model across 4 devices, varying the model dimension, and simulate training steps for this shallow model while measuring optimizer step time. We report the average over the final 20 steps out of 100 total training steps.
Communication costs are identical between Muon and \textsc{Dion2} in this setup.

Even in this limited setting, the left plot in \autoref{fig:main} shows a clear reduction in optimizer step time as the selection fraction decreases, demonstrating the computational benefit of orthonormalizing smaller submatrices. When communication savings are also accounted for, the overall benefits are even larger.

\section{Conclusion}

To improve the scalability of Muon, this work proposes a simple method to reduce the size of the matrix that needs to be orthogonalized. The success of sparse updates, even with a randomly selected submatrix, is surprising, and it would be interesting to investigate whether these sparse updates provide additional benefits. The empirical evaluations presented here are still preliminary, and more thorough experiments are required. In particular, although \textsc{Dion2} can offer substantial advantages over Muon in terms of computation and communication costs, large-scale experiments are needed to fully assess these benefits. We leave these directions to future work.
 
\bibliographystyle{plainnat}
\bibliography{refs}

% \newpage
% \appendix

% \renewcommand{\appendixpagename}{\centering \LARGE Appendix}
% \appendixpage

% \startcontents[section]
% \printcontents[section]{l}{1}{\setcounter{tocdepth}{2}}
% \newpage

%%%%%%%%%%%%%%%%%%%%%%%%%%%%%%%%%%%%%%%%%%%%%%%%%%%%%%%%%%%%
\end{document}